\documentclass[sigconf]{acmart}
\AtBeginDocument{%
  }

\setcopyright{acmlicensed}
\copyrightyear{2018}
\acmYear{2018}
\acmDOI{XXXXXXX.XXXXXXX}
\acmConference[Conference acronym 'XX]{conference title}{June 03--05,
  2018}{Woodstock, NY}
\acmISBN{978-1-4503-XXXX-X/2018/06}

\usepackage{graphicx}     
\usepackage{subcaption}   

\usepackage{algorithm} 
\usepackage{algpseudocode} 

\usepackage{multirow}
\usepackage{graphicx}

\algrenewcommand\algorithmicrequire{\textbf{Input:}}
\algrenewcommand\algorithmicensure{\textbf{Output:}}

\begin{document}

\title{Semi-supervised Instruction Tuning for Large Language Models on Text-Attributed Graphs}

\author{Zixing Song}
\email{zixing.song@bristol.ac.uk}
\orcid{0000-0002-8871-3990}
\affiliation{%
  \institution{University of Bristol}
  \city{Bristol}
  \country{United Kingdom}
}

\author{Irwin King}
\email{king@cse.cuhk.edu.hk}
\orcid{0000-0001-8106-6447}
\affiliation{%
  \institution{The Chinese University of Hong Kong}
  \city{New Territories}
  \country{Hong Kong SAR}}

\renewcommand{\shortauthors}{Song et al.}

\begin{abstract}
The emergent reasoning capabilities of Large Language Models (LLMs) offer a transformative paradigm for analyzing text-attributed graphs—the fundamental data structures underpinning social networks and digital communities. However, deploying LLMs for social good applications, such as detecting misinformation or moderating toxic communities, is frequently hindered by the scarcity of high-quality annotations. While instruction tuning is the prevailing method for adapting pre-trained LLMs to graph learning tasks like node classification, it requires a substantial volume of annotated (INSTRUCTION, OUTPUT) pairs deriving from labeled nodes. This requirement is particularly prohibitive in the social domain, where obtaining expert labels for sensitive or evolving content is costly and slow. Furthermore, standard graph instruction tuning fails to exploit the vast amount of unlabeled nodes, which contain latent correlations due to edge connections that are beneficial for downstream predictions. To bridge this gap, we propose a novel \textit{Semi-supervised Instruction Tuning pipeline for Graph Learning}, named \textit{SIT-Graph}. Notably, SIT-Graph is model-agnostic and can be seamlessly integrated into any graph instruction tuning method that utilizes LLMs as the predictor. SIT-Graph operates via an iterative self-training process. Initially, the model is fine-tuned using instruction pairs constructed solely from the labeled nodes. Then it generates confidence-filtered pseudo-responses for unlabeled nodes to strategically augment the dataset for the next round of fine-tuning. Finally, this iterative refinement progressively aligns the LLM with the underlying node correlations. Extensive experiments demonstrate that when incorporated into state-of-the-art graph instruction tuning methods, SIT-Graph significantly enhances their performance on text-attributed graph benchmarks, achieving over 20\% improvement under the low label ratio settings.
\end{abstract}


\maketitle

\section{Introduction}
\label{sec:intro}
\begin{figure}[!t]
    \centering
    \begin{subfigure}[b]{\linewidth}
        \centering
        \includegraphics[width=\linewidth]{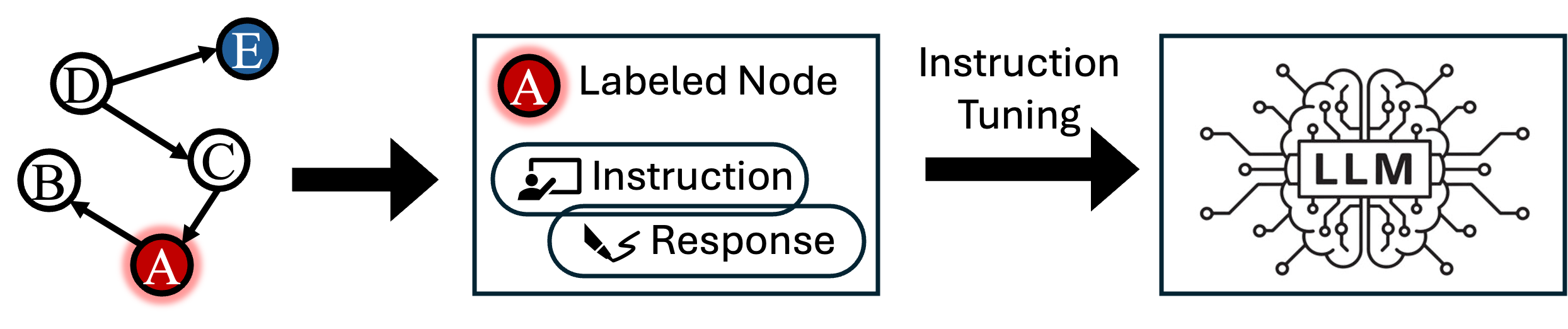}
        \caption{Existing instruction tuning for graph learning tasks.}
        \label{fig:previous}
    \end{subfigure}

    \begin{subfigure}[b]{\linewidth}
        \centering
        \includegraphics[width=\linewidth]{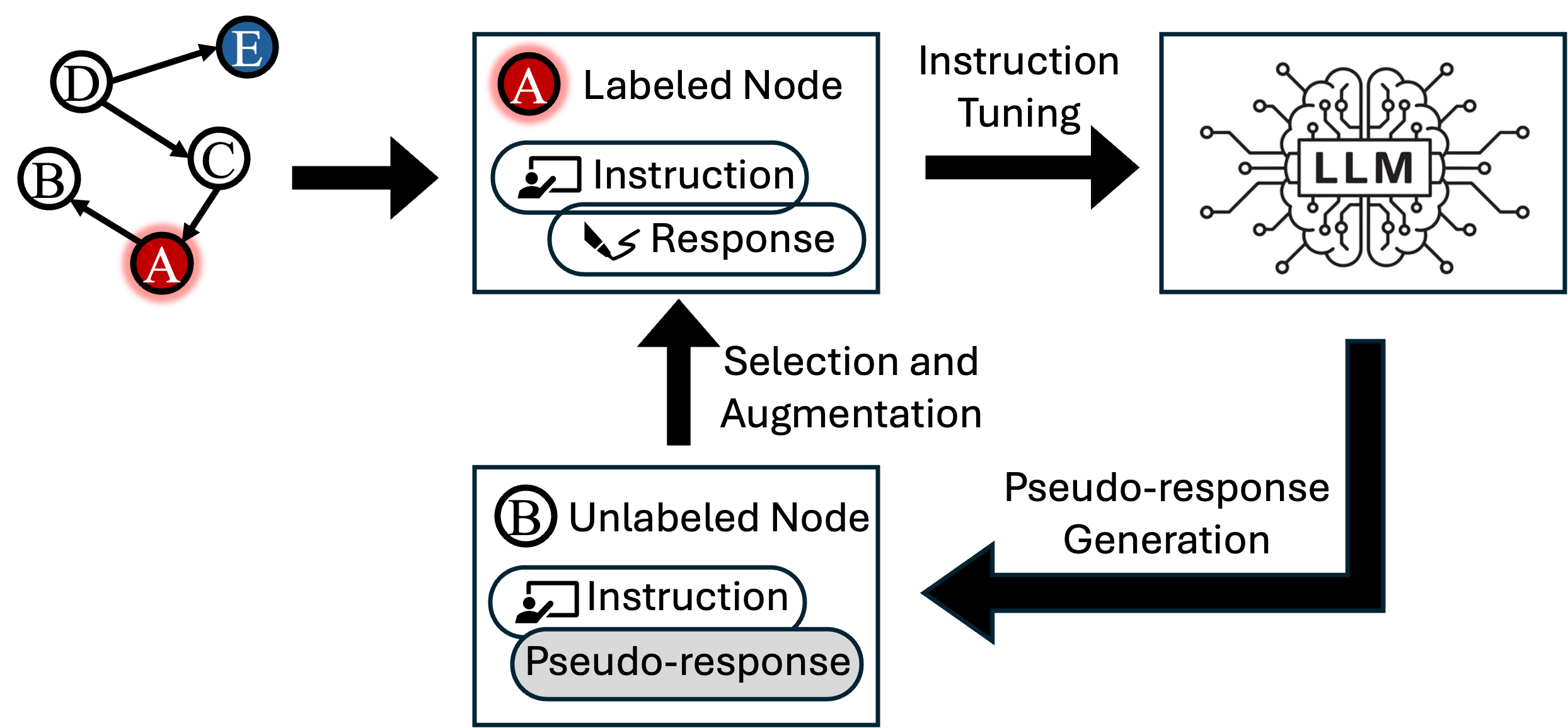}
        \caption{Semi-supervised instruction tuning for graph learning tasks.}
        \label{fig:ours}
    \end{subfigure}

    \caption{Comparison of existing IT methods (a) and ours (b).}
    \label{fig:intro}
\end{figure}

Modern digital ecosystems, from the World Wide Web to complex social networks, are intrinsically modeled as text-attributed graphs~\cite{DBLP:conf/www/Zhang0YL24}. In this multimodal data structure, nodes encompass rich semantic content (e.g., a web page's text or a user's post) while the edges define the topological context (e.g., hyperlinks or social follows). This dual data modality presents a unique challenge for automated content analysis: effective reasoning requires the simultaneous processing of textual semantics and structural information.

For instance, in phishing website classification~\cite{DBLP:journals/corr/abs-2305-05378}, a malicious webpage may perfectly mimic the textual content of a legitimate banking site, easily deceiving a text-only classifier. However, its structural context—such as a lack of inbound links from trusted domains or isolation from the dense core of the web graph—often reveals its true fraudulent nature. Thus, for social good applications on the World Wide Web, relying solely on text is insufficient, and the model must scrutinize the interplay between the textual content of the nodes and the structural connectivity in the graph.

This fundamental challenge—fusing node-level rich text with graph-level relational topology—has motivated the application of Large Language Models (LLMs) to text-attributed graphs~\cite{GraphCLIP,DBLP:conf/kdd/FangFZT24}. A promising direction involves hybrid methods that employ Graph Neural Networks (GNNs) to explicitly encode the topological structure. The objective is to extract topological information into representations that LLMs can leverage, thereby enhancing the LLM's powerful semantic reasoning with the GNN's structural awareness.

Instruction tuning (IT) is the prevailing paradigm to adapt LLMs for graph learning tasks~\cite{InstructGraph,MuseGraph,GraphGPT}, such as node classification. This supervised fine-tuning (SFT) process aligns the LLM by fine-tuning it on a dataset of task-specific handcrafted (INSTRUCTION, OUTPUT) pairs~\cite{DBLP:journals/jair/ZhangWDZTC25}. Critically, current methods construct these handcrafted pairs exclusively from the labeled nodes (Figure~\ref{fig:previous}). This dependency creates a bottleneck, as the high cost of annotation on text-attributed graphs results in a severely limited number of labeled nodes, fundamentally constraining the adaptation.

This reliance on only labeled nodes ignores the rich structural correlations in unlabeled nodes. Nodes are connected and the neighbors of unlabeled nodes may offer predictive signals. For example, an unlabeled website heavily interlinked with a known phishing site is itself highly suspicious. This observation leads to the central question: \textit{How can we conduct instruction tuning for LLMs on text-attributed graphs in a semi-supervised manner, effectively leveraging both limited labeled nodes and the vast unlabeled nodes?} (Figure~\ref{fig:ours})

Designing a semi-supervised instruction tuning pipeline is non-trivial. First, classical semi-supervised learning (SSL) is designed for deterministic models~\cite{DBLP:conf/iclr/KipfW17}, not generative LLMs that output free-form text. This necessitates a new pseudo-response framework integrating graph structural priors. Second, quantifying the confidence of a response is notoriously difficult~\cite{luo-etal-2025-semi}. A flawed metric risks catastrophic error propagation on graphs, as incorrect label prediction can be propagated via edge connections. Finally, the pipeline must ensure self-evolution~\cite{DBLP:conf/aaai/TanCZXD25} so that the LLM genuinely learns from the unlabeled nodes instead of amplifying its own biases.

To address these challenges, we are the first to propose a novel \textbf{S}emi-supervised \textbf{I}nstruction \textbf{T}uning pipeline for generative LLMs to solve \textbf{Graph} learning tasks (\textbf{SIT-Graph}). Our framework operates via an iterative self-training process. Initially, the model is fine-tuned using instruction pairs derived solely from the limited labeled nodes, learning to map the aligned graph tokens and text tokens to the target handcrafted response. Subsequently, the tuned model generates pseudo-responses for unlabeled nodes. To mitigate error propagation, we introduce a robust filtering mechanism to quantify the confidence of these pseudo-responses. Finally, only high-quality pseudo-responses are added to the instruction dataset for the next round of fine-tuning. This self-evolution refinement allows the model to progressively learn from unlabeled nodes, preventing the amplification of the bias in the pre-trained LLM.

In summary, our contributions are threefold. 
\begin{itemize}
    \item We propose SIT-Graph, a novel semi-supervised framework that is model-agnostic and seamlessly integrates into any LLM-based graph instruction tuning methods.
    \item We design an iterative self-training mechanism that leverages confidence-filtered pseudo-responses for unlabeled nodes to ensure model self-evolution.
    \item We demonstrate that SIT-Graph significantly improves standard supervised graph instruction tuning baselines.
\end{itemize}

\section{Related Work}
\label{sec:related}
\subsection{Graph Neural Networks} 
Graph Neural Networks (GNNs) have become a powerful tool for learning representations from text-attributed graphs~\cite{DBLP:conf/www/Zhang0YL24,DBLP:conf/nips/SongZK23}. GNNs excel in capturing the intricate relationships within graph topology. GNNs learn node representations through repeated message propagation and aggregation operations. The learned node representations can be tailored for graph learning tasks like node classification.

\subsection{Large Language Models for Graph Learning} 
Large Language Models (LLMs) for graph learning is an emerging area that primarily follows two paradigms for integrating GNNs and LLMs on text-attributed graphs. The first employs GNNs as the primary predictor~\cite{DBLP:conf/iclr/ChenMWH0Z0T24}, with LLM-enhanced features as input. However, these models typically lack generalizability across diverse datasets. The second leverages LLMs as the primary predictor~\cite{GraphGPT,MuseGraph}, adapting them via instruction tuning to achieve generalizability. The drawback is their reliance on a large, fully-annotated corpus of instruction-response pairs. In contrast, the proposed SIT-Graph addresses this bottleneck by introducing a novel semi-supervised instruction tuning pipeline.

\section{Methodlogy}
\label{sec:method}

\subsection{Problem Definition}
Let $\mathcal{G}=({V},{E},\mathbf{A},\mathbf{X}, C)$ be a text-attributed graph. The set of $N$ nodes is partitioned into labeled $V_L$ and unlabeled $V_U$ sets ($V = V_L \cup V_U$). The graph structure is represented by the adjacency matrix $\mathbf{A} \in \mathbb{R}^{N\times N}$. Node attributes are provided as a feature matrix $\mathbf{X} \in \mathbb{R}^ {N \times F}$ (containing numerical/categorical features) and a collection of raw text documents $C = \{c_i\}_{i=1}^N$. Given the observed labels $y_i$ for all $v_i \in V_L$, the objective is to predict the labels $\tilde{y}_i$ for all nodes $v_i \in V_U$.

\begin{figure*}
    \centering
    \includegraphics[width=0.92\linewidth]{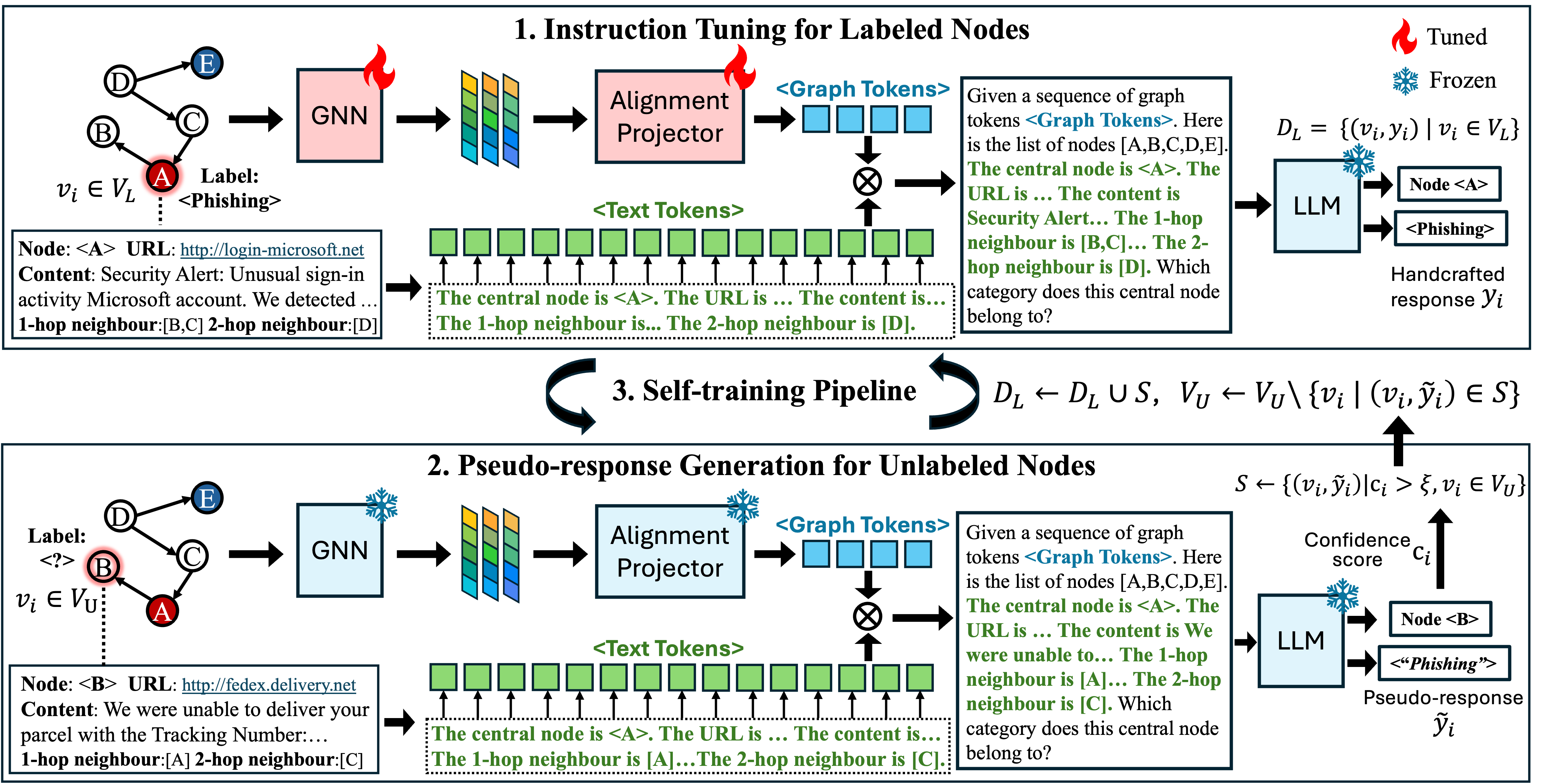}
    \caption{Semi-supervised instruction tuning pipeline for graph learning (SIT-Graph). First, it conducts instruction tuning, adapting a pre-trained LLM to map combined graph tokens and text tokens to the handcrafted response for labeled nodes (Sec.~\ref{sec:it}). Second, the tuned model generates pseudo-responses and confidence scores for unlabeled nodes (Sec.~\ref{sec:pseudo_gen}). Finally, high-quality pseudo-responses are added to the augmented dataset, and the model evolves in a self-training manner (Sec.~\ref{sec:self-training}).}
    \label{fig:model}
\end{figure*}

\subsection{Instruction Tuning for Labeled Nodes}
\label{sec:it}
In this first step of the proposed SIT-Graph model (Figure~\ref{fig:model}), we conduct instruction tuning on labeled nodes $v_i \in V_L$ in the standard supervised fashion. The objective is to adapt the pre-trained LLM to capture both the node-level textual information and the graph-level topological structure from the given text-attributed graph.

To encode graph structure information into the LLMs, SIT-Graph generates specialized graph tokens, rather than using rule-based graph linearization methods (i.e., natural language prompts). This process begins by using a backbone GNN to extract node embeddings, $\mathbf{H} = \mathbf{H}^L$, via the following message-passing updates:
\begin{equation}
    \mathbf{H}^{l} = \sigma \left(\tilde{\mathbf{A}}\mathbf{H}^{l-1}\Theta_g\right), \quad 1\leq l \leq L.
\end{equation}
Here, $\sigma(\cdot)$ is the activation function. $\tilde{\mathbf{A}} = \mathbf{A}+\mathbf{I}$ is the self-loop adjacency matrix. $L$ is the total number of layers in the GNN and $H^0 = \mathbf{X}$. $\Theta_g$ is the GNN model parameter.

To enhance the alignment between graph structure and textual information, we employ an alignment projector $\Theta_p$, which can be as simple as a multilayer perceptron (MLP). This projector maps the GNN node representations $\mathbf{H}$ into a sequence of graph tokens. These graph tokens are designed to align within the LLM's embedding space with the text tokens extracted from the node-level content $C$. The aligned graph tokens create a modified token sequence for the LLM, denoted as $\{<\texttt{graph}\_\texttt{token}>_1, \cdots, <\texttt{graph}\_\texttt{token}>_N\}$, which corresponds to the number of $N$ nodes in the graph. 

For each labeled node $v_i \in V_L$, the final instruction input for the LLM is constructed by combining the graph tokens and its text tokens. The corresponding target response $y_i$ is handcrafted from the ground-truth label. To optimize the tuning process efficiently, using the available labeled nodes, we fine-tune only the GNN and projector components while keeping the LLM parameters frozen.

\subsection{Pseudo-response Generation for Unlabeled Nodes}
\label{sec:pseudo_gen}
After fine-tuning the SIT-Graph model with the labeled nodes, we utilize it to generate the pseudo-response for unlabeled nodes. These new (instruction, response) pairs are then used to augment the original labeled dataset, creating a new dataset for retraining the model. By fixing the parameters of the SIT-Graph model, we obtain the pseudo-response $\tilde{y}_i$ as the prediction for each unlabeled node $v_i \in V_U$. However, a key challenge is that these pseudo-responses may contain noise or low-quality predictions, which could potentially misguide the retraining process.

To address this issue, we filter the generated pseudo-responses to select only high-quality input-output pairs for unlabeled nodes. We use the model's confidence in its output $\tilde{y}_i$ as our selection criterion. As the LLM generates the response token by token, we approximate this confidence-based entropy by calculating the per-token negative log-likelihood. For each unlabeled node $v_i \in V_U$, the entropy $H(\tilde{y}_i)$ is computed on pseudo-response $\tilde{y}_i$ as
\begin{equation}
    H(\tilde{y}_i) = - \frac{1}{L_i} \sum_{k=1}^{L_i}\log P(\tilde{y}_i^k \mid v_i, \tilde{y}_i^{<k}),
\end{equation}
where $L_i$ is the length of the pseudo-response $\tilde{y}$ generated by the SIT-Graph model, $\tilde{y}_i^k$ is the $k$-th token in the response, and $\tilde{y}_i^{<k} = \{\tilde{y}_i^1,\tilde{y}_i^2,\cdots,\tilde{y}_i^{k-1}\}$ are the preceding tokens. We obtain the output confidence 
\begin{equation}
    c_i = 1 - H(\tilde{y}_i).
\end{equation}
We set a threshold $\xi$ and select confident responses with the corresponding unlabeled nodes as the new dataset $S$ for retraining.
\begin{equation}
    S \gets \{(v_i, \tilde{y}_i) \mid c_i > \xi, v_i \in V_U\}.
\end{equation}

\subsection{Self-training Pipeline}
\label{sec:self-training}

\setlength{\textfloatsep}{8pt}
\begin{algorithm}[t]
\caption{Self-Training Pipeline for SIT-Graph}
\label{alg:self_training}
\begin{algorithmic}[1]
\Require 
    Initial SIT-Graph model with GNN $\Theta_g$ and Projector $\Theta_p$);
    Original labeled dataset $D_L = \{(v_i, y_i) | v_i \in V_L\}$; 
    Unlabeled node set $V_U$; 
    Confidence threshold $\xi$; 
    Number of iterations $T$.
\State $S \gets \emptyset$.
\For{$t = 1$ \textbf{to} $T$ \textbf{and} $V_U \neq \emptyset$}
\State Fine-tune $\Theta_g$ and $\Theta_p$ on $D_L$. \Comment{Sec.~\ref{sec:it}}
\ForAll{node $v_i \in V_U$} \Comment{Sec.~\ref{sec:pseudo_gen}}
        \State Generate pseudo-response $\tilde{y}_i$ with $\Theta_g$, $\Theta_p$ frozen.
        \State Calculate confidence score $c_i$.
    \EndFor
    \State $S \gets \{(v_i, \tilde{y}_i) \mid c_i > \xi, v_i \in V_U\}$.
    \State $D_L \gets D_L \cup S$, $V_U \gets V_U \setminus \{v_i \mid (v_i, \tilde{y}_i) \in S\}$.
\EndFor
\State \Return The fine-tuned model with $\Theta_g$ and $\Theta_p$.
\end{algorithmic}
\end{algorithm}

Finally, we combine the selected pseudo-response dataset $S$ with the original input-output dataset (i.e., $D_L \gets D_L \cup S$) and further perform the instruction tuning on this augmented dataset once again. This self-training process can be iterated for several rounds to progressively enhance the model's adaptability for graph learning tasks. By leveraging both the high-quality pseudo-response for unlabeled nodes and the ground-truth handcrafted responses for labeled nodes, we significantly improve the label efficiency of the model. The complete self-training pipeline of the proposed SIT-Graph model is detailed in Algorithm 1. Notably, SIT-Graph is model-agnostic and can be seamlessly integrated into any graph instruction tuning methods that utilize LLMs as the predictor.

\section{Experiments}
\label{sec:experiment}
\subsection{Setup} 
We evaluate Graph-SIT with three datasets: Cora, arxiv (two citation networks), and Wiki-CS (a web link graph). The statistics of the datasets and the default data splits can be found in~\cite{DBLP:conf/nips/0001WZCJ0C024}.

\begin{table}[t]
\centering
\caption{Relative performance improvement of graph instruction tuning baselines when integrated with SIT-Graph.}
\label{tab:main_results}
\resizebox{\columnwidth}{!}{%
\begin{tabular}{clccccc}
\toprule
 &  & \multicolumn{3}{c}{\textbf{Node Classification}} & \multicolumn{2}{c}{\textbf{Link Prediction}} \\ 
\cmidrule(lr){3-5} \cmidrule(lr){6-7}
\textbf{Model} & \textbf{Setting} & Cora & Wiki-CS & arxiv & Cora & Wiki-CS \\ 
\midrule

\multirow{3}{*}{\shortstack{GraphGPT \\ \cite{GraphGPT}}}
 & Vanilla & 23.25 & 6.30 & 21.45 & 64.23 & 67.32 \\
 & w/ SIT-Graph & \textbf{37.42} & \textbf{9.88} & \textbf{34.15} & \textbf{84.91} & \textbf{86.45} \\
 & \textit{Improvement} & \small{(+61.0\%)} & \small{(+56.8\%)} & \small{(+59.2\%)} & \small{(+32.2\%)} & \small{(+28.4\%)} \\
\midrule

\multirow{3}{*}{\shortstack{InstructGraph \\ \cite{InstructGraph}}}
 & Vanilla & 89.33 & 76.46 & 81.50 & 89.89 & 94.72 \\
 & w/ SIT-Graph & \textbf{90.15} & \textbf{85.88} & \textbf{88.74} & \textbf{90.34} & \textbf{95.88} \\
 & \textit{Improvement} & \small{(+0.9\%)} & \small{(+12.3\%)} & \small{(+8.9\%)} & \small{(+0.5\%)} & \small{(+1.2\%)} \\
\midrule

\multirow{3}{*}{\shortstack{LLaGA \\ \cite{LLaGA}}}
 & Vanilla & 74.42 & 73.88 & 72.78 & 86.82 & 90.54 \\
 & w/ SIT-Graph & \textbf{88.95} & \textbf{80.12} & \textbf{87.65} & \textbf{89.45} & \textbf{93.08} \\
 & \textit{Improvement} & \small{(+19.5\%)} & \small{(+8.4\%)} & \small{(+20.4\%)} & \small{(+3.0\%)} & \small{(+2.8\%)} \\
\midrule

\multirow{3}{*}{\shortstack{InstructGLM \\ \cite{InstructGLM}}}
 & Vanilla & 69.10 & 45.73 & 39.09 & 76.11 & 86.45 \\
 & w/ SIT-Graph & \textbf{85.34} & \textbf{65.21} & \textbf{60.88} & \textbf{90.05} & \textbf{92.14} \\
 & \textit{Improvement} & \small{(+23.5\%)} & \small{(+42.6\%)} & \small{(+55.7\%)} & \small{(+18.3\%)} & \small{(+6.6\%)} \\
\midrule

\multirow{3}{*}{\shortstack{MuseGraph \\ \cite{MuseGraph}}}
 & Vanilla & 71.86 & 65.72 & 63.14 & 79.37 & 88.83 \\
 & w/ SIT-Graph & \textbf{86.44} & \textbf{81.95} & \textbf{78.56} & \textbf{88.75} & \textbf{93.62} \\
 & \textit{Improvement} & \small{(+20.3\%)} & \small{(+24.7\%)} & \small{(+24.4\%)} & \small{(+11.8\%)} & \small{(+5.4\%)} \\
\midrule

\multirow{3}{*}{\shortstack{GraphCLIP \\ \cite{GraphCLIP}}}
 & Vanilla & 67.31 & 70.19 & 65.85 & 83.15 & 92.67 \\
 & w/ SIT-Graph & \textbf{84.15} & \textbf{85.06} & \textbf{82.91} & \textbf{90.22} & \textbf{94.85} \\
 & \textit{Improvement} & \small{(+25.0\%)} & \small{(+21.2\%)} & \small{(+25.9\%)} & \small{(+8.5\%)} & \small{(+2.4\%)} \\
\bottomrule
\end{tabular}%
}
\end{table}

For baselines, we choose several popular graph instruction tuning models, including GraphGPT~\cite{GraphGPT}, InstructGraph~\cite{InstructGraph}, LLaGA~\cite{LLaGA}, InstructGLM~\cite{InstructGLM}, MuseGraph~\cite{MuseGraph}, and GraphCLIP~\cite{GraphCLIP}, which all use the LLM as the final predictor to solve the graph learning tasks.

\subsection{Main Results}
As SIT-Graph is model-agnostic, it functions as a universal plug-and-play module capable of enhancing any existing graph instruction tuning method. To empirically validate this adaptability, we integrate SIT-Graph into six representative state-of-the-art baselines. We evaluate the performance gain across three benchmark datasets on both node classification and link prediction tasks. We fix the ratio of labeled to unlabeled nodes at 1\%.  The results, summarized in Table~\ref{tab:main_results}, demonstrate that SIT-Graph consistently delivers substantial performance improvements across all baselines and tasks. This confirms that the inclusion of pseudo-response signals from unlabeled nodes provides a generalizable benefit that is not tied to a specific graph instruction tuning method.

\subsection{Impact of Label Ratio}
We further investigate the performance of SIT-Graph by varying the ratio of labeled to unlabeled nodes with the backbone fixed as GraphCLIP. Figure 3 demonstrates that the performance gain excels under severe low label ratio settings (under $\approx 10^{-2}$), achieving 20–30\% improvement. As the labeled node set approaches the size of the unlabeled node set, the performance gap naturally narrows, validating SIT-Graph's superior label efficiency in scenarios with minimal supervision.

\subsection{Ablation Study}
\begin{figure}[t]
    \centering
    \begin{minipage}[c]{0.5\linewidth}
        \centering
        \includegraphics[width=\linewidth]{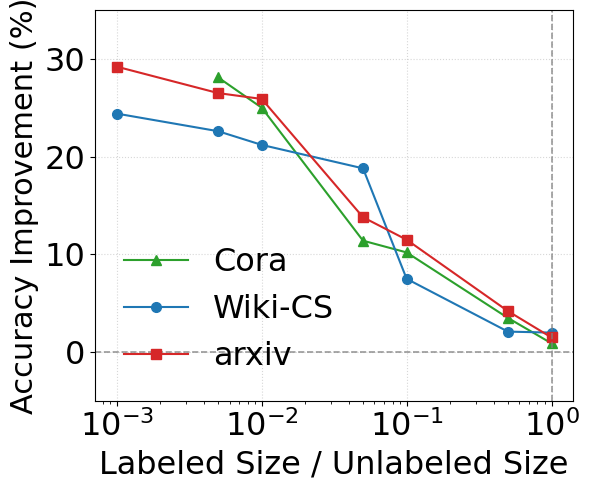}
        \captionof{figure}{Impact of the labeled-to-unlabeled node ratio on accuracy improvement.}
        \label{fig:improvement_trend}
    \end{minipage}
    \hfill 
    \begin{minipage}[c]{0.47\linewidth}
        \centering
        \captionof{table}{Ablation study of the SIT-Graph model for node classification on Cora. AP stands for the alignment projector. CF stands for the confidence filter.}
        \label{tab:ablation_study}
        
        \resizebox{\linewidth}{!}{%
            \begin{tabular}{cc}
            \toprule
            \textbf{Model Variant} & \textbf{Accuracy (\%)} \\ 
            \midrule
            \textbf{SIT-Graph} & \textbf{84.15} \\ 
            \midrule
            \hspace{2mm} w/o GNN & 77.04  \\
            \hspace{2mm} w/o AP & 72.69 \\ 
            \hspace{2mm} w/o CF & 80.23 \\
            \bottomrule
            \end{tabular}%
        }
    \end{minipage}
\end{figure}

We examine the contribution of each component using the Cora dataset and the GraphCLIP as the backbone. In Table~\ref{tab:ablation_study}, the Alignment Projector proves most critical, followed by the GNN, confirming the need for aligned graph tokens. Additionally, robust filtering is vital to prevent noisy pseudo-labels from poisoning the self-training loop.

\section{Conclusion}
\label{sec:conclusion}
In this work, we presented \textsc{SIT-Graph}, a semi-supervised instruction tuning framework designed to overcome the critical bottleneck of label scarcity in analyzing text-attributed graphs. By effectively harnessing the multimodal synergy between node-level textual semantics and graph-level topology, our approach offers a robust solution for content analysis in data-sparse environments. This efficiency is potentially vital for social good applications—ranging from detecting emerging misinformation to monitoring community dynamics—where obtaining expert annotations is often resource-prohibitive. Ultimately, \textsc{SIT-Graph} paves the way for more accessible, scalable, and structure-aware LLM systems capable of effectively safeguarding the integrity of digital ecosystems.

\bibliographystyle{ACM-Reference-Format}
\bibliography{reference}

\end{document}